\pdfoutput=1
\documentclass[11pt]{article}

\usepackage[]{ACL2023}
\usepackage{times}
\usepackage{latexsym}
\usepackage{graphicx}
\usepackage{textcomp}

\usepackage[T1]{fontenc}
\usepackage[utf8]{inputenc}

\usepackage{microtype}

\usepackage{inconsolata}

\usepackage{booktabs, makecell, tabularx, array}
\newcolumntype{C}{>{\centering\arraybackslash}X} 
\newcolumntype{D}{>{\arraybackslash}X} 
\usepackage{multirow}

\usepackage{amsmath}

\usepackage{verbatim}

\usepackage{xspace}

\newcommand{\dataset}{{{\textsc{Model-Personas}}}\xspace}

\newcommand{\camerareadytext}[1]{#1 \xspace}

\definecolor{lightergray}{RGB}{230, 230, 230}

\title{You don't need a personality test to know these models are unreliable: Assessing the Reliability of Large Language Models on Psychometric Instruments}

\author{
  Bangzhao Shu$^{\dagger}$\thanks{~~Equal contribution}  ~ Lechen Zhang$^{\dagger}$\footnotemark[1]  ~ Minje Choi$^{\ddagger}$  ~ Lavinia Dunagan$^{\dagger}$ ~ Lajanugen Logeswaran$^\sharp$ \\
  \textbf{Moontae Lee}$^\sharp$ ~ \textbf{Dallas Card}$^{\dagger}$ ~ \textbf{David Jurgens}$^{\dagger}$ \\
$^\dagger$University of Michigan, Ann Arbor, MI, USA  \\
$^\ddagger$Georgia Institute of Technology, Atlanta, GA, USA  \\
$^\sharp$LG AI Research, Ann Arbor, MI, USA  \\
$^{\dagger}${ \tt \{bangzhao, leczhang, laviniad, dalc, jurgens\}@umich.edu}\\
$^\ddagger${ \tt minje.choi@gatech.edu} ~ $^\sharp${ \tt \{llajan, moontae.lee\}@lgresearch.ai}
}

\begin{document}
\maketitle
\begin{abstract}
The versatility of Large Language Models (LLMs) on natural language understanding tasks has made them popular for research in social sciences. To properly understand the properties and innate \textit{personas} of LLMs, researchers have performed studies that involve using prompts in the form of questions that ask LLMs about particular opinions. In this study, we take a cautionary step back and examine whether the current format of prompting LLMs elicits responses in a consistent and robust manner. We first construct a dataset that contains 693 questions encompassing 39 different instruments of persona measurement on 115 persona axes. Additionally, we design a set of prompts containing minor variations and examine LLMs' capabilities to generate answers, as well as prompt variations to examine their consistency with respect to content-level variations such as switching the order of response options or negating the statement. Our experiments on 17 different LLMs reveal that even simple perturbations significantly downgrade a model's question-answering ability, and that most LLMs have low negation consistency. Our results suggest that the currently widespread practice of prompting is insufficient to accurately and reliably capture model perceptions, and we therefore discuss potential alternatives to improve these issues.
\end{abstract}

\section{Introduction}
\label{sec:intro}
Large Language Models (LLMs), trained on a massive and diverse volume of human-generated text corpora, show remarkable capabilities in carrying out instruction-based tasks and achieving high performance on several NLP benchmarks~\cite{brown2020gpt3,touvron2023Llama2,taori2023alpaca}. Notably, LLMs possess the capability to produce coherent text based on complex instructions, a feature that has paved the way for their application in the development of conversational assistants and chatbots.
This advancement has encouraged research into the extent to which these models exhibit characteristics similar to human cognition and behavior, leading to several studies that focus on measuring the psychological properties or the \textit{persona} of LLMs using specifically designed prompts. Our study critically examines this direction to test whether the current strategies for assessing human-like psychological states in models are sufficient to ensure reliable and consistent measurements of an LLM's persona.

For humans, a persona encompasses a broad class of attributes that make up a person's identity, such as personality, demographics, or values, which all influence how people portray themselves \cite{cheng2023marked}. This terminology has been adopted in several NLP studies which range from identifying personas from text corpora~\cite{bamman2013learning,chu2018learning,ghosh2022empersona,zhu2023paed} to injecting personas into language generation tasks~\cite{ahn2023mpchat,xu2022long,lee2022personachatgen,li2023learning}.
As LLMs are increasingly used in interpersonal settings, it is beneficial to have accurate measurements of latent properties in the model that can influence what text they generate in order to mitigate any potential harm that may arise from undesired innate model biases~\cite{lucy2021gender,feng2023pretraining}.
As a result, several recent studies have investigated the tendencies of LLMs such as ChatGPT from angles such as political preferences~\cite{liu2022political}, personality tests~\cite{pan2023mbti,vganesan2023systematic,miotto2022gpt,jiang2023evaluating}, and moral choices~\cite{santurkar2023whose,cheng2023marked}.

Current approaches to measuring dimensions of LLMs' personas typically assess them like humans, by turning the questions in psychological instruments into prompts and scoring the answers~\cite{serapiogarcía2023personality}. Although models are specifically trained to answer questions in general, multiple works have raised concerns about the brittleness of this capability~\cite{sclar2024quantifying}, pointing out, for example, their sensitivity to prompt formats.  Further, recent studies have shown that LLMs struggle with questions that contain cues such as negation and thus generate inconsistent results rather than fully comprehending the question~\cite{garcíaferrero2023negation}. Thus, we investigate the behavior of LLMs in generating responses to persona-related questionnaires from three angles: (1) \textbf{Comprehensibility}: are LLMs capable of understanding instructions and generating answers given a specific prompt? (2) \textbf{Sensitivity}: do model answers vary with spurious changes to the question format? (3) \textbf{Consistency}: do model answers vary with different content-level changes to the question?

This study makes the following three contributions.
First, we curate \dataset, a large panel of 39 psychological instruments, and standardize these into 693 questions across 115 axes.
Second, we introduce a systematic evaluation framework for testing the sensitivity and consistency of LLMs' answers to persona questions through controlled variations of the prompts.
Third, we evaluate multiple open-source LLMs using \dataset, showing that models vary widely across the sensitivity and consistency levels with most models having no consistent persona. Our results reveal that BLOOMZ models are most robust to sensitivity perturbations, while FLAN-T5 models are most consistent. In general, however, most LLMs failed to deliver robust answers, raising concerns about the validity of claims with respect to models' ``personalities'' or ``values.''

\section{\dataset: A Comprehensive Benchmark for Measuring Personas}
\label{sec:persona}

Studies for creating and identifying personas largely involve qualitative methods such as interviews, field studies, and surveys~\cite{brickey2012comparing,salminen2020persona}. In particular, surveys in the form of questionnaires have widely been adopted in psychology and behavioral studies to measure personality traits and opinions of individuals in a standardized manner at a large scale~\cite{spence1974personal,dalbert1999world,patrick2002development,van2000multicultural}. These questionnaires, known as psychological instruments, are frequently calibrated through experiments to capture to core axes of variation in people.
Questionnaires are easily compatible with LLMs pretrained with instruction-based prompts, as these models can provide a wide range of outputs  ranging from open-ended responses to simple yes/no answers. Further, prompting is widely accepted as the default method for eliciting responses from LLMs. \camerareadytext{Following this trend, our benchmark also takes the form of a questionnaire designed to prompt an answer in a yes/no format.}

In constructing a benchmark for assessing model persona, we performed a comprehensive survey of existing instruments. The selection criteria were focused on mostly stable persona attributes, and excluded instruments focusing on mental health.
Our persona instruments can be categorized into five groups: Belief statements, Normative statements, Values, Descriptors, and Situations. Belief statements include instruments that reflect an individual's conviction about the truth of a particular idea, such as Unjust World Scale (UWS)~\cite{DALBERT2001561} and Money Attitudes Measure (MAM)~\cite{Furnham2020ANM}; Normative statements include instruments that express value judgments, opinions, or prescriptions about how things ought to be, such as Holistic Cognition Scale (HCS)~\cite{HCS} and Ambivalent Classism Inventory (ACI)~\cite{doi:10.1080/01973533.2020.1828084}; Values include instruments that examine an individual's deeply held beliefs about what is important or desirable and serve as guiding principles for behavior and judgment, such as Strength Based Inventory (SBI)~\cite{doi:10.1080/13674676.2020.1744310}; Descriptors include instruments that are used to detail personality traits, such as Big 5 Personality Traits (OCEAN)~\cite{Poropat2009AMO}; Situation includes instruments that measure individuals' responses and behaviors in various social contexts and scenarios, such as Emotional Response to Unfairness Scale (ERUS)~\cite{BIZER2020109882}.

Each instrument contains one or more axes that evaluate a specific dimension. For example, the Ethics Position Questionnaire (EPQ)~\cite{forsyth1980taxonomy} contains two axes: Idealism and Relativism, which evaluate an individual's ethical position from two different aspects. Furthermore, each axis contains one or more statements, and individuals can get a score on an axis based on how strongly they agree or disagree with the statements.
Overall, our dataset consists of 693 questions in English under 39 instrument categories and 115 axes, encapsulating a broad spectrum of psychological and sociological constructs. The sample instruments are shown in Appendix Table \ref{table:persona-instrument-examples}.

Instruments each have their own question format or phrasing, which introduces undesirable variability when evaluating LLMs. Therefore, across all instruments, we introduce a standard question format to prompt models with. Following best practice on prompt design (e.g.,\citet{aher2023using}), we use a structured prompt of "\texttt{Statement:\textbackslash n<Statement>\textbackslash nQuestion:\textbackslash nDo you agree with the statement? Reply with only `Yes' or `No' without explaining your reasoning.\textbackslash nAnswer:\textbackslash n}", and then generate one token from the model to get an answer. This prompt is designed to elicit assent or dissent with the question's premise.  
Recognizing that models vary in their ability to understand negation~\cite{jang2023negated,garcíaferrero2023negation}, during standardization, we rephrase questions with any explicit negation such that the intent is the same but the negation is removed. 
This paraphrasing allows us to systematically introduce negation later to test the model's answering consistency.

\section{Design of Prompt Variants}
\label{sec:prompt}

Given that LLMs can be sensitive to the format~\cite{sclar2024quantifying} and content~\cite{min2023recent} of prompts, here, we introduce the design choices for perturbing the prompts. These changes are intended to affect the comprehensibility, sensitivity, and consistency of an LLM's answer for a given instrument question.

\subsection{Prompts for Spurious Variation}
Our first analysis centers on whether spurious changes to the input prompt can affect model predictions when inferring persona. Here, spurious changes refer to subtle adjustments to the prompt that leave the question content unchanged. Such perturbations, theoretically, should not alter the model's confidence in generating an answer, as the semantic meaning of the sentence remains unchanged. Four types of prompt variations are used:

\noindent \textbf{Sentence Ending}: We compare two types of sentence ending: "\texttt{?}" and "\texttt{:}". An example would be "\texttt{Your Answer?}" versus "\texttt{Your Answer:}".

\noindent \textbf{Colon+<\textbackslash s>}: We test whether varying the number of spaces after the colon by adding zero spaces, one space, double space, or a line-break can affect performance. For example, does "\texttt{Answer: }" produce different results from "\texttt{Answer:\textbackslash n}"?

\noindent \textbf{Answer/Response}: We compare the use of the word used at the end of the prompt: "\texttt{Answer:}" or "\texttt{Response:}".

\noindent \textbf{Section Separation Format}: We compare different formats to separate sections (Statement/Question/Answer) in our prompt. The separators include Line-break, Single Space, Double-Bar (//) and Triple-Sharp (\#\#\#).

\noindent
Full examples can be found in Appendix Table \ref{table:prompt-variant-examples}.

\subsection{Prompts for Content-level Variation}
Even if LLMs are able to understand the instructions and generate a valid answer with high confidence, it is possible that they are merely generating based on the question structure rather than on their understanding of the question. To contrast with the spurious variations, we construct a set of perturbations targeting the question content to examine whether LLMs can generate consistent responses. Four types of prompts are used:

\noindent \textbf{Option Consistency}: We test the consistency of response when asked to return different types of labels. For example, the responses of an LLM when asked to answer "\texttt{Reply with only `Yes' or `No'}" should be consistent with being asked to answer "\texttt{Reply with only `True' or `False'}".

\noindent \textbf{Negation Consistency}: LLM predictions are known to be affected by the inclusion of negation words~\cite{jang2023negated,garcíaferrero2023negation}. We test this by manually rewriting each question into reversed meaning and looking at the changes in response. We test two types of negation: 
(1) \textbf{Direct Negation}: We insert a negation word such as ``not'', ``no'', or ``don't'' in syntactically coherent position to reverse the answer's polarity.
(2) \textbf{Paraphrastic Negation}: We reverse the meaning of the sentence by rephrasing it without including a negation word.
Examples of this are in Appendix Table~\ref{table:negation-examples}. 
    
\noindent \textbf{Order Consistency}: We test the consistency of model generations when the given response options are in reversed order. For example, if we ask LLMs to answer using "\texttt{Reply with only `Yes' or `No'}", the answer should be consistent with being asked to answer using "\texttt{Reply with only `No’ or `Yes’}".

\section{Experimental Setup}

Here, we describe the experimental setup and define the metrics for evaluating model performance.

\subsection{Measuring Model Comprehensibility}
We define a model's \textit{comprehensibility} as the ability to generate an answer corresponding to one of the available options, e.g., ``True'' or ``False''. Therefore, we calculate the proportion of answers whose first token is valid. For each question $q$'s response $R(q)$, it is considered valid if $R(q) \in P \cup N $, where $P$ and $N$ are the set of possible valid positive and negative answers to the prompt's question.

Therefore, the model $M$'s comprehensibility can be defined as:
$
\textrm{Com}(M) = \frac{\# R(q) \in (P \cup N)}{\# R(q)}
$ for all questions $q$ in \dataset.

\subsection{Measuring Sensitivity and Consistency}

If a model can answer the prompt, to what degree do its answers vary when the format and content of the question are varied?  We define a model's \textit{sensitivity} as the degree to which its answers change when prompted with spurious variations, and a model's \textit{consistency} as the degree to which a model agrees across different paraphrases of the same question.

For each question $q$ in the instrument dataset $D$, we first measure the model's response $R(q)$ as the valid response option with the highest probability. 
We then modify $q$ into a different prompt $q'$. This modification can either occur as a spurious change~(\S\ref{sec:prompt}.1) or at content-level~(\S\ref{sec:prompt}.2). We then obtain $R(q')$ as well.

Since LLMs should generate answers that are robust to perturbations, we measure both sensitivity and consistency as the fraction of samples from which the answers did not change after perturbation. However, for negation consistency, we expect the model to answer with the reverse option to be consistent with the non-negated original prompt; negation consistency is measured as the number of opposite answers for $q'$ relative to the answer for $q$. 

\subsection{Comparison with Psychometric Measurements of Consistency and Reliability}

\camerareadytext{
Given a person's answers to a psychometric instrument, prior work in Psychology has examined whether these answers are internally consistent---i.e., is the person answering at random or do the relationships between answers indicate the stable presence of some construct. Such consistency and reliability scores are measured through metrics like  Cronbach's $\alpha$~\cite{cronbach1951coefficient}, Guttman's $\lambda_6$~\cite{guttman1945basis}, and McDonald's $\omega$~\cite{mcdonald2013test}. Recent work has examined using these methods in  case studies for measuring the personalities of LLMs using psychometric instruments such as HEXACO~\cite{miotto2022gpt} or Big Five Inventory~\cite{serapiogarcía2023personality}. Especially in the case of \citet{serapiogarcía2023personality}, the authors show that LLMs contain personality traits, which are both reliable and valid across several of these metrics when prompted with multiple questions, suggesting that, collectively, the answers are self-consistent with each other. 
}

\camerareadytext{
In contrast to studies of \textit{inter}-question consistency, our study focuses on a related question about \textit{intra}-question consistency: If the same question was asked in a slightly different way, would the answer change? Thus the two approaches provide complementary information. Our approach builds on recent work that tests whether (or how) prompting a model with two versions of an input to assess whether the model can generate the same output~\cite[e.g.,][]{webson2022prompt,sclar2024quantifying}. 
Here, the consistency is not across items as in the case of \citet{serapiogarcía2023personality}, but rather within item. 
Our study starts with the expectation that an answer should be the same in these within-item tests---i.e., a human would answer the question the same way, regardless of whether the question was phrased as true/false vs. yes/no. 
Therefore, we measure consistency as the percentage of samples that reach the same answer regardless of perturbations.}

\begin{table*}
\setlength\tabcolsep{1.5pt}
\fontsize{7}{8}\selectfont
\begin{tabularx}{\textwidth}{@{} r *{18}{c} @{}}
\toprule
\multicolumn{1}{c}{\multirow{3}{*}{\normalsize\textbf{Model}}}
&\multicolumn{1}{c}{\textbf{falcon}}
&\multicolumn{1}{c}{\textbf{RedPajama}}
&\multicolumn{4}{c}{\textbf{BLOOMZ}}
&\multicolumn{4}{c}{\textbf{Llama2}}
&\multicolumn{4}{c}{\textbf{FLAN-T5}} 
&\multicolumn{3}{c}{\textbf{GPT}}
&\multicolumn{1}{c}{\multirow{3}{*}{\textbf{Average}}}
\\

\cmidrule(r){2-2} \cmidrule(r){3-3} \cmidrule(r){4-7} \cmidrule(r){8-11} \cmidrule(r){12-15} \cmidrule{16-18}

&7B
&7B-Instruct
&560M &1B1 & 3B &7B1
&7B &7B-Chat &13B &13B-Chat
&Small &Base &Large &XL
&GPT-2 &GPT-3.5 &GPT-4 
\\
\midrule
\textbf{Colon Ending*} & 1.00 & \multicolumn{1}{c}{0.07} & 1.00 & 1.00 & 1.00 & 1.00 & 1.00 & 1.00 & 1.00 & 1.00 & 1.00 & 1.00 & 1.00 & 1.00 & 0.01 & 1.00 & 0.98 & 0.89\\ 
\textbf{Question-Mark Ending} & 1.00 & \multicolumn{1}{c}{0.00} & 1.00 & 1.00 & 1.00 & 1.00 & 0.00 & 0.94 & 0.03 & 0.30 & 1.00 & 1.00 & 1.00 & 1.00 & 0.00 & 0.99 & 0.95 & 0.65\\ 
\midrule
\textbf{Colon + Line-Break*} & 1.00 & \multicolumn{1}{c}{0.01} & 1.00 & 1.00 & 1.00 & 1.00 & 1.00 & 1.00 & 1.00 & 1.00 & 1.00 & 1.00 & 1.00 & 1.00 & 0.01 & 1.00 & 0.98 & 0.89\\ 
\textbf{Colon + No Space} & 1.00 & \multicolumn{1}{c}{0.83} & 1.00 & 1.00 & 1.00 & 1.00 & 0.91 & 1.00 & 0.56 & 1.00 & 1.00 & 1.00 & 1.00 & 1.00 & 0.01 & 1.00 & 0.99 & 0.90\\ 
\textbf{Colon + Single Space} & 0.00 & \multicolumn{1}{c}{0.00} & 1.00 & 1.00 & 1.00 & 1.00 & 0.00 & 1.00 & 0.00 & 1.00 & 1.00 & 1.00 & 1.00 & 1.00 & 0.00 & 1.00 & 0.98 & 0.70\\ 
\textbf{Colon + Double Space} & 0.00 & \multicolumn{1}{c}{0.24} & 1.00 & 1.00 & 1.00 & 1.00 & 0.94 & 1.00 & 0.97 & 1.00 & 1.00 & 1.00 & 1.00 & 1.00 & 0.00 & 1.00 & 0.99 & 0.82\\
\midrule
\textbf{Answer:*} & 1.00 & \multicolumn{1}{c}{0.01} & 1.00 & 1.00 & 1.00 & 1.00 & 1.00 & 1.00 & 1.00 & 1.00 & 1.00 & 1.00 & 1.00 & 1.00 & 0.01 & 1.00 & 0.98 & 0.89\\ 
\textbf{Response:} & 0.96 & \multicolumn{1}{c}{0.01} & 1.00 & 1.00 & 1.00 & 1.00 & 0.24 & 1.00 & 1.00 & 1.00 & 1.00 & 1.00 & 1.00 & 1.00 & 0.00 & 1.00 & 0.97 & 0.83\\ 
\midrule
\textbf{Line-Break Separated*} & 1.00 & \multicolumn{1}{c}{0.01} & 1.00 & 1.00 & 1.00 & 1.00 & 1.00 & 1.00 & 1.00 & 1.00 & 1.00 & 1.00 & 1.00 & 1.00 & 0.01 & 1.00 & 0.98 & 0.89\\ 
\textbf{Single Space Separated} & 0.93 & \multicolumn{1}{c}{0.60} & 1.00 & 1.00 & 1.00 & 1.00 & 1.00 & 1.00 & 1.00 & 1.00 & 1.00 & 1.00 & 1.00 & 1.00 & 0.01 & 0.99 & 1.00 & 0.92\\ 
\textbf{Double-Bar Separated} & 0.83 & \multicolumn{1}{c}{1.00} & 1.00 & 1.00 & 1.00 & 1.00 & 1.00 & 1.00 & 1.00 & 1.00 & 0.76 & 1.00 & 1.00 & 1.00 & 0.07 & 1.00 & 1.00 & 0.92\\ 
\textbf{Triple-Sharp Separated} & 1.00 & \multicolumn{1}{c}{0.18} & 1.00 & 1.00 & 1.00 & 1.00 & 1.00 & 1.00 & 1.00 & 1.00 & 1.00 & 1.00 & 1.00 & 1.00 & 0.08 & 1.00 & 0.99 & 0.90\\

\bottomrule
\end{tabularx}
\caption{Model's Comprehensibility of Different Prompt Variants. Baseline format options are marked with an asterisk (*). We discovered that different prompt formats can cause a huge difference in the model's comprehensibility.}
\label{table:model-comprehensibility-result}
\end{table*}

\begin{table*}[t]
\setlength\tabcolsep{3.5pt}
\fontsize{7}{8}\selectfont
\begin{tabularx}{\textwidth}{@{} r c c c cccc cccc cccc cc c @{}}
\toprule
\multicolumn{1}{c}{\multirow{3}{*}{\normalsize\textbf{Model}}}
&\multicolumn{1}{c}{\textbf{falcon}}
&\multicolumn{4}{c}{\textbf{BLOOMZ}}
&\multicolumn{4}{c}{\textbf{Llama2}}
&\multicolumn{4}{c}{\textbf{FLAN-T5}} 
&\multicolumn{2}{c}{\textbf{GPT}} 
&\multicolumn{1}{c}{\multirow{3}{*}{\textbf{Average}}}
\\

\cmidrule(r){2-2} \cmidrule(r){3-6} \cmidrule(r){7-10} \cmidrule(r){11-14} \cmidrule{15-16}

&7B
&560M &1B1 & 3B &7B1
&7B &7B-Chat &13B &13B-Chat
&Small &Base &Large &XL
&GPT-3.5 & GPT-4
\\
\midrule
\textbf{Question-Mark Ending} & 0.86 & 0.46 & 0.95 & 0.73 & 0.63 & 0.00 & 0.33 & 0.02 & 0.08 & 0.80 & 0.91 & 0.93 & 0.96 & 0.84 & 0.88 & 0.60\\ 
\midrule
\textbf{Colon + No Space} & 0.56 & 0.47 & 0.95 & 0.91 & 0.59 & 0.81 & 0.80 & 0.34 & 0.86 & 0.98 & 0.98 & 0.99 & 0.99 & 0.94 & 0.94 & 0.81\\ 
\textbf{Colon + Single Space} & 0.00 & 0.66 & 0.95 & 0.86 & 0.48 & 0.00 & 0.77 & 0.00 & 0.84 & 1.00 & 1.00 & 1.00 & 1.00 & 0.92 & 0.93 & 0.69\\ 
\textbf{Colon + Double Space} & 0.00 & 0.43 & 0.95 & 0.86 & 0.56 & 0.55 & 0.71 & 0.37 & 0.86 & 1.00 & 1.00 & 1.00 & 1.00 & 0.93 & 0.93 & 0.74\\
\midrule
\textbf{Response:} & 0.79 & 0.98 & 0.97 & 0.99 & 0.93 & 0.17 & 0.67 & 0.54 & 0.83 & 0.88 & 0.97 & 0.97 & 0.99 & 0.97 & 0.95 & 0.84\\ 
\midrule
\textbf{Single Space Separated} & 0.48 & 0.44 & 0.95 & 0.92 & 0.57 & 0.75 & 0.77 & 0.55 & 0.89 & 0.98 & 0.98 & 0.99 & 0.99 & 0.91 & 0.94 & 0.81\\ 
\textbf{Double-Bar Separated} & 0.49 & 0.36 & 0.95 & 0.94 & 0.57 & 0.89 & 0.66 & 0.33 & 0.83 & 0.32 & 0.77 & 0.91 & 0.96 & 0.92 & 0.93 & 0.72\\ 
\textbf{Triple-Sharp Separated} & 0.87 & 0.90 & 0.97 & 0.95 & 0.77 & 0.48 & 0.85 & 0.82 & 0.95 & 0.96 & 0.92 & 0.92 & 0.98 & 0.93 & 0.91 & 0.88\\

\bottomrule
\end{tabularx}
\caption{Model's sensitivity to different prompt variations relative to the baseline format shows that most LLMs' responses are sensitive to trivial changes, except for the Flan-T5 family.}
\label{table:sensitivity-results}
\end{table*}

\subsection{Model Details}
Using our consistency metrics, we perform evaluations on several variants of open-source LLMs which are widely used in current research. The models included in our experiments are GPT-2~\cite{radford2019gpt2}, Falcon-7B~\cite{refinedweb2023falcon}, BLOOMZ (560M, 1B1, 3B, 7B1)~\cite{muennighoff2022BLOOMZ}, Llama2 (7B, 7B-Chat, 13B, 13B-Chat)~\cite{touvron2023Llama2}, RedPajama-7B~\cite{together2023redpajama}, and FLAN-T5 (Small, Base, Large, XL)~\cite{chung2023flant5}. We also included the results from closed-source LLMs such as GPT-3.5 and GPT-4~\cite{openai2023gpt4} in our consistency test. The temperature was set to 0.0 for all experiments to minimize the effects of randomness. Additional model inference details are reported in Appendix \ref{sec:inferencedetail}.

\section{Results}
Here, we present our results on the robustness of LLM predictions on \dataset.

\begin{figure*}[t]
    \centering
    \includegraphics[width=0.92\textwidth]{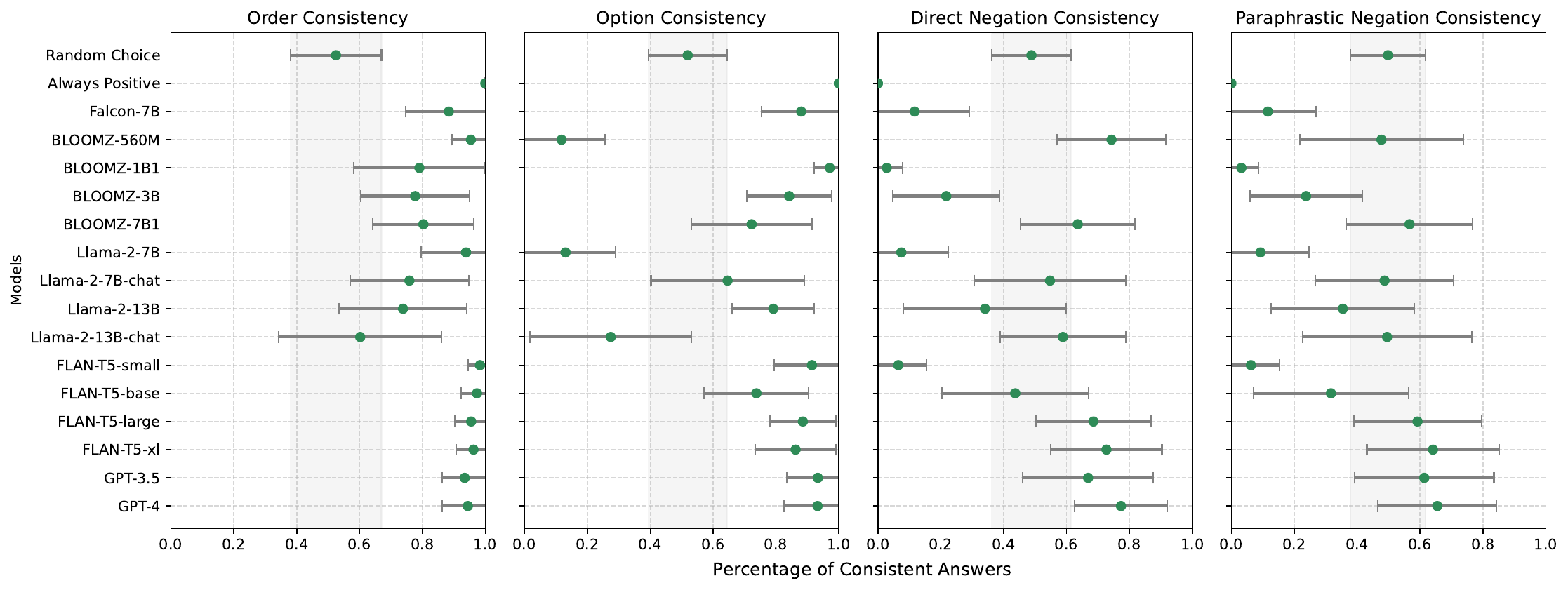}
    \caption{A comparison of LLMs on different consistency metrics. The area shaded in \colorbox{lightergray}{gray} indicates the consistency of answering with a random valid response. We discover that while most LLMs provide consistent results regarding order and option consistency, they struggle with both cases of negation consistency.}
    \label{fig:model_consistency}
\end{figure*}

\subsection{LLMs differ in Comprehensibility}

Models varied widely in their ability to generate a valid answer to the instruments' questions, as shown in Table~\ref{table:model-comprehensibility-result}.
Models from the BLOOMZ and FLAN-T5 families demonstrate a uniformly high likelihood of responding correctly to all variations of the prompts. In evaluations using nine varied prompt formats, the BLOOMZ family models return valid responses to all prompts. The FLAN-T5 models also respond correctly to most of the variations of the prompts, except FLAN-T5 small to Double-Bar Separated format.
Falcon-7B, RedPajama-7B, Llama 2-7B, and Llama 2-13B show varied performance when faced with different prompt formats. For instance, in Falcon-7B, adding a single space after a colon can drastically cut the comprehensibility score from 1.0 to 0.0, indicating that its ability to respond as ``True'' or ``False'' to a given question is harshly impeded.

Our results suggest that psychological instruments cannot be blindly given to models without first testing whether the model will cooperate with the prompts. Subtle changes in prompt syntax can significantly influence the performance of some models in validly answering questions, which, depending on how an experimenter handles non-answers, may significantly influence the model's scores on the instrument.

\subsection{LLMs can be Sensitive even to Spurious Prompt Variation}

Even when models can validly answer questions, our experiments show that their answers may change due to small, spurious differences in the format of the prompt itself. We examine the sensitivity of the models with relatively high comprehensibility (we exclude GPT-2 and RedPajama, which shows poor comprehensibility among most of the prompt variants). Table~\ref{table:sensitivity-results} shows the sensitivity of each LLM in comparison with the baseline prompt setting (which is marked with an asterisk in Table~\ref{table:model-comprehensibility-result}). Ideally, an LLM should not change their answer when asked the same question with slightly different prompt formats, especially under trivial changes such as changing a single space to a double space or line break. Nevertheless, we observe that several LLMs change their responses when prompted using such variations. In several cases, we observe that the sensitivity score of a model in a particular setting is similar to random (0.5), though it is hard to find a consistent pattern among the cases where models express sensitivity. Notably, Most LLMs in the FLAN-T5 family (Base, Large, and XL) exhibit perfect robustness to most of the perturbations. Despite LLMs of the BLOOMZ family constantly being the most comprehensible of the instructions across prompt variations~(Table~\ref{table:model-comprehensibility-result}), BLOOMZ-560M's and BLOOMZ-7B1's answers change frequently, nearing the consistency of random behavior. This experiment suggests that while possibly correlated, being able to return answers of high confidence does not entail robustness to sensitivity and vice versa.

\subsection{Staying Consistent is Challenging for LLMs}

We now turn to see whether LLMs are capable of understanding the persona questions and providing consistent answers that suggest a latent persona property. We examine the four types of consistencies of 15 previous models with high comprehensibility, including GPT-3.5 and GPT-4~\cite{openai2023gpt4}. Figure~\ref{fig:model_consistency} shows the different consistencies of models, and below we summarize the trends.

\paragraph{Most LLMs maintain Order Consistency and Option Consistency}

All models show Order Consistency performance above random choice. Most models scored over 0.7, indicating moderate consistency. No clear relationship emerges between model size and order consistency, and performance disparities are also present across different model families, with FLAN-T5 models and GPT models leading. It is important to note that a high order consistency score does not unequivocally indicate model superiority, as models that consistently respond positively---regardless of prompt---will naturally score higher.
The variance in option consistency is also noteworthy. Within model families, larger models do not always outperform their smaller counterparts, though Llama2-13B does outperform Llama2-7B. When models of the same size from different families are compared, the performance varies.
Similar to order consistency, a higher score in option consistency does not necessarily mean the model is performing well; it could indicate a tendency to respond positively regardless of the prompt. BLOOMZ-1B1, for instance, shows high option consistency but low negation consistency, suggesting it provides uniform answers independent of prompt variations. BLOOMZ-560M exhibits lower option consistency, indicating a potential disparity in its performance on True/False versus Yes/No questions. The FLAN-T5 model family and GPT models stand out for their stability and superior performance in option consistency.

\paragraph{Negation Consistency is hard to achieve}

While most LLMs maintain consistency levels over a random-answering baseline for order consistency and option consistency, \textit{all} models struggle to generate consistent answers when the meaning of the question is reversed using negation, either with the direct inclusion of negative words or through semantic changes. These results align with recent work on negation prompts~\cite{garcíaferrero2023negation} which showed that understanding negation is challenging for various LLMs. In fact, the majority of models (10 of 15) achieve a score close to random (0.5) or even worse regarding negation consistency. Only five models exhibit higher consistency on both direct negation and paraphrastic negation dimensions—primarily among larger models including FLAN-T5-Large, FLAN-T5-XL, BLOOMZ-7B, GPT-3.5 and GPT-4. Interestingly, BLOOMZ-560M, the smallest of the BLOOMZ family reaches high direct negation consistency. 
It can also be seen that models tend to achieve higher consistency when direct negation words are used rather than the sentence being semantically negated. We can observe from Figure \ref{fig:model_consistency} that all models perform worse on paraphrastic negations than on direct negations. A potential reason is that paraphrastic negation introduces subtle shifts and requires a deeper understanding of the context to be able to provide a flipped answer, whereas for direct negation the negation word itself can lead to a flipped answer.

\paragraph{Summary}

In summary, there is a significant consistency variation in the performance of the tested models, with larger models generally exhibiting a greater likelihood of consistent responses across the four metrics examined. Nevertheless, the majority fail to outperform a simple random choice. Notably, the BLOOMZ-560m model displays exceptional consistency with True/False questions but significantly less so with Yes/No questions. The FLAN-T5 family consistently performs well across all metrics of persona consistency. \camerareadytext{We also display the consistency scores for each axis averaged across the different models in Appendix Figures \ref{fig:consistency_by_axes_1} and \ref{fig:consistency_by_axes_2}, which further highlight significant variation across models even when tested on the same axis.}

The key implication of this result is that a simple prompting of a model with an instrument's questions is \textit{not} sufficient to claim any persona. Instead, models must be prompted with at least negated forms of the questions to verify the model's answers indicate a deeper understanding of the prompt and not just an artifact.

For our purposes, we consider a model to exhibit consistent personas if it achieves a threshold score of 0.6 for the four consistencies, which was selected via manual inspection. Of the 15 models evaluated, only three---Flan-T5-XL, GPT-3.5 and GPT-4, met this criterion, suggesting the potential for these models to possess consistent personas. Flan-T5-Large, with Paraphrastic Negation Consistency slightly lower than 0.6, almost satisfy the requirement.

\section{Can Adding Personas Improve Consistency?} 
\label{sec:adding_personas}

Most LLMs achieve low consistency scores when tested on prompt variations. However, most LLMs are not explicitly design to behave as a ``person'' and so may not have an implicit tendency to respond consistently like a person would. Commonly, models are prompted with a persona to have it embody a certain personality. Thus, we examine whether explicitly adding details of a specific persona in a prompt can enable the model to produce more consistent results.

\begin{figure*}[t]
    \centering
    \includegraphics[width=\textwidth]{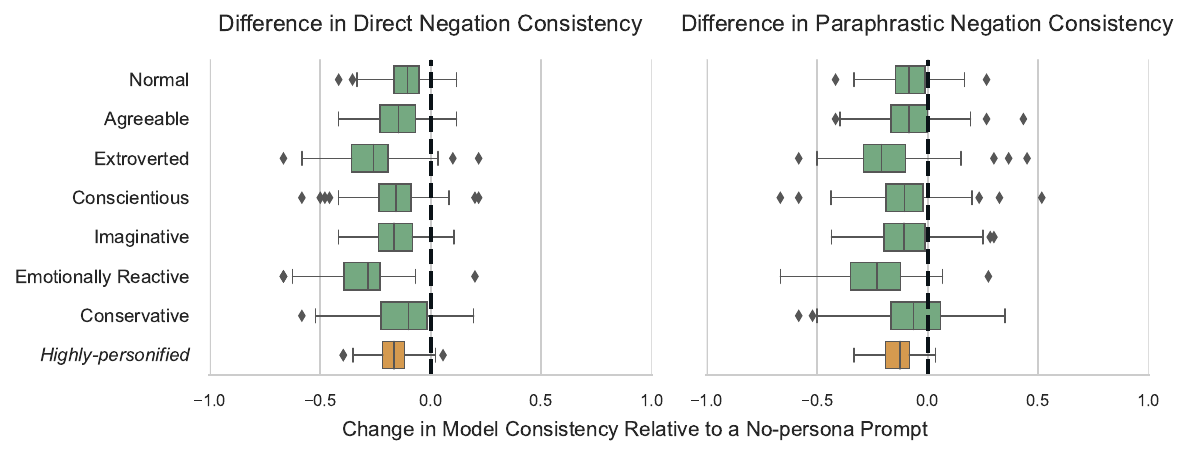}
    \caption{Negation Consistency Shift after adding specific personalities into the prompt. Adding personalities decreases the general negation consistency of LLMs, even if some axes' consistencies are increased as outliers.}
    \label{fig:consistency_shift_persona_inject}
\end{figure*}

\paragraph{Experimental Setup}
To test whether adding a persona can improve model consistency, we first obtained predictions under various settings: (1) \textbf{Baseline setting}: The prompt does not contain any persona and is the same as in the previous section. (2) \textbf{Normal person}: All questions begin with ``You are a normal person'' at the start of the prompt, aiming to guide the model to adopt the perspective of an individual without any additional information biasing a response towards one or more personality attributes. (3) \textbf{Specific personality}: we explicitly mention the type of personality in the prompt level along with a brief description of the personality type (e.g. ``You are an extrovert who is outgoing, sociable, and energized by interactions with other people.''). The full set of specific personality prompts can be found in Appendix Table \ref{table:prompts_for_personas}. (4) \textbf{Highly-personified}: our final setting corresponds to a prompt containing all of a curated list of 35 different personalities characteristics in an attempt to constrain the model outputs on all instruments (see Appendix Table \ref{table:prompts_for_personas}). The motivation for this final design is to test whether specifying a large number attributes related to multiple personality attributes will improve consistency across most of the dimensions.

We obtain the consistency scores for all instruments across 10 different models. Once obtained, we compute the \textbf{consistency shift}, which depicts the change in consistency with respect to the baseline setting (1). For a particular model and instrument, the consistency shift is obtained by subtracting the consistency score under the baseline setting from the adjusted prompt. By averaging across all models and all instruments, we obtain the final consistency shift.

\paragraph{Results}

The level of consistency shift on both types of negation under different prompt variations is shown in Figure~\ref{fig:consistency_shift_persona_inject}. We observe that adding \textit{any} personality to the prompt \textit{decreases} the general negation consistency of LLMs, which is the same even for the ``Normal Person'' prompt that does not hint at any personality. However, this drop does not occur uniformly across all instruments, as can be seen in the box plots with values greater than 0. 

Through manual inspection, we discover several cases in which the axes where consistency improves are relevant to the personality that is being injected into the prompt. For example, instruments measuring extroversion increased in consistency when prompted using an extroverted personality in the prompt. 
This attribute specific improvement suggests that consistency could perhaps be improved by adding multiple descriptions in the persona to ensure all attributes relate in some way. However, we observe that our Highly Personified setting that contains such descriptions is among the least consistent. 
Overall, our results suggest that while adding more personality information at a prompt level can improve the consistency of relevant dimensions, this gain may be shadowed by consistency drop in several unrelated dimensions, and that adding multiple types of personality information does not help.

Together, our results  suggesting injecting a specific persona into a prompt to generate consistent outputs has limited benefits, at best, and LLMs with such personas as a part of their system prompt should not be expected to be more consistent.

\begin{figure*}[t]
    \centering
    \includegraphics[width=0.9\textwidth]{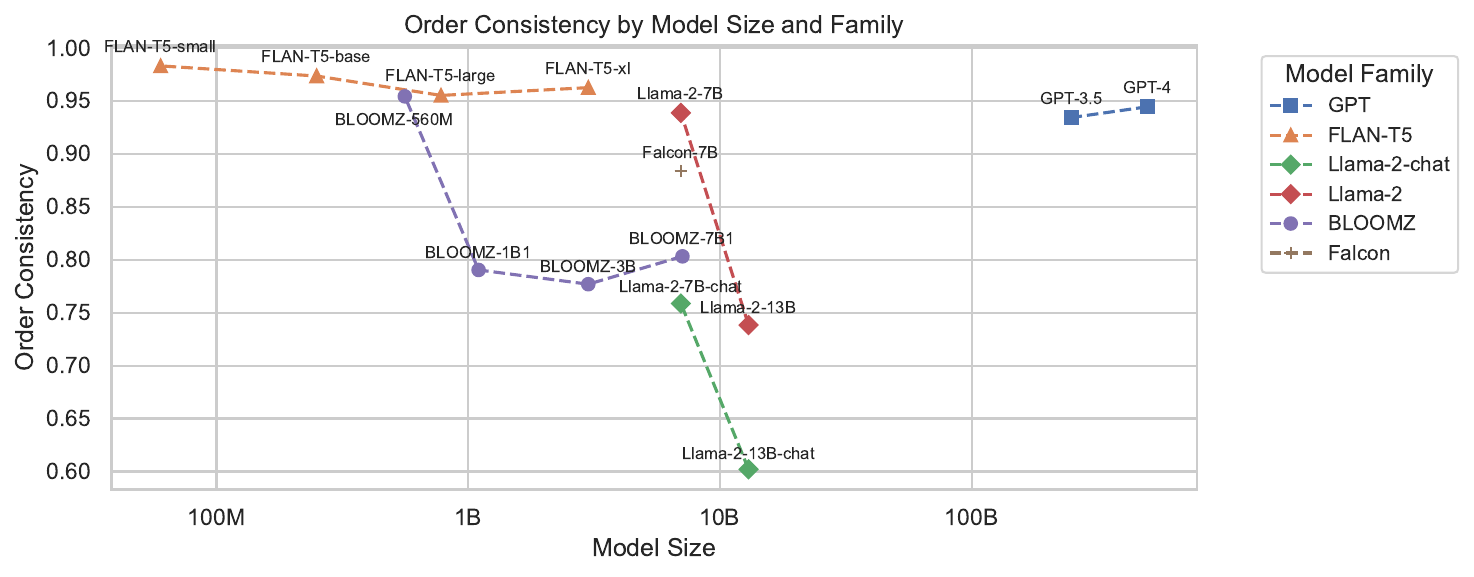}
    \caption{A comparison of model size and consistency when changing the order of the answers.}
    \label{fig:consistency_lineplot3}
\end{figure*}

\section{Discussion}
In this section, we discuss the implications of our study as well as future steps for addressing and mitigating inconsistency issues.

\paragraph{Sensitivity and Inconsistency of LLMs Question their Measurement Capabilities}
Despite the rapidly increasing view of LLMs as a means of understanding and emulating human responses across various fields of social sciences,  our results show that most models fail to generate consistent responses even when tested on simple variants of input prompts. This calls into question whether the predictions generated by LLMs in response to probes on social constructs such as moral decisions, public opinions, or political ideologies can be truly seen as valid. The unreliability and inconsistency of current LLMs can pose a challenge for practitioners who plan to conduct tasks based on the personalities of these models.

\paragraph{Mitigating the Unreliability of Prompts}
What measures can we adopt to mitigate the current unreliability of LLMs? We offer two suggestions based on prior studies. One approach would be to perform a preliminary test on the confidence scores of the answers for a given set of prompts before running the prompts to obtain the preferences towards each persona. For instance, \citet{aher2023using} propose selecting one of $k$  prompts choices to maximize the validity rate, then conducting subsequent experiments on that prompt.

Another promising approach is to perform additional fine-tuning steps to improve model robustness. However, performing additional fine-tuning steps might not always be beneficial. For instance, a recent study has shown that even after additionally fine-tuning LLMs on text corpora that include negation samples does not significantly improve its capability to understand negation \citep{garcíaferrero2023negation}. Besides, additional fine-tuning might alter the LLM's innate persona that was present before fine-tuning, which raises an issue in reproducibility and generalizability.

\noindent
\textbf{Model Size vs. Architecture Type}
Interestingly, our results indicate that the reliability of LLMs is not necessarily correlated with a model's number of parameters, which is consistent with recent findings indicating that larger model sizes do not always lead to higher task performance or task understandability~\cite{choi2023llms}. For example, Figure \ref{fig:consistency_lineplot3} shows that models do not become more consistent when varying the order of the options as the number of parameters increase; Appendix Figures \ref{fig:consistency_lineplot1}, \ref{fig:consistency_lineplot2}, and \ref{fig:consistency_lineplot4} show similar results for the three other consistency measurements. Rather, we observe that comprehensibility, sensitivity, and consistency scores can be better grouped at the architecture family level. This trend was particularly notable for the models belonging to the FLAN-T5 and BLOOMZ families, showing that design details in the pertaining phase might have a profound effect on an LLM's zero-shot capabilities when prompted to provide answers in downstream tasks.

\section{Conclusion}

Human-like interactions with large language models can inspire a desire to assess models like humans. In the psychological setting, this can mean assessing whether models have human-like persona traits, such as personality or values, using questionnaires as prompts. However, does the text models generate in response reflect a consistent latent attribute of the model---or just a continuation of a high probability sequence?  

Here, we systematically assess whether LLMs are capable of generating robust responses for assessing personas by evaluating the extent to which LLMs can understand questions and provide answers in a consistent manner under various prompt variations. 
Our evaluations on the \dataset dataset suggest that  
the answers given by most widely-used LLMs are not consistent with any latent persona attributes and instead are, in part, driven by features of the prompt. Not only do models vary in their ability to generate a valid answer, relative to spurious changes in format, but the answers themselves---e.g., whether a model affirms an extroversion preference---are also sensitive to such changes. Furthermore, most LLMs fail to deliver consistent preferences when the question meaning is reversed using negation. In fact, only one (Flan-T5-XL) of the fifteen open-source models, and two closed-source GPT models, achieved a reasonable average consistency score over 0.6. 

Overall, our study demonstrates the unreliability of a blind application of a psychological questionnaire for assessing the attributes of LLMs, and calls for cautionary measures such as sensitivity and consistency checks to ensure robustness of measurement.
The code and dataset are available at \url{https://github.com/orange0629/llm-personas}.

\section{Limitations}
Our study is not without its limitations.
The first is that, apart from the proprietary GPT models, we only experimented on LLMs of small to medium sizes. Despite studies showing greater capabilities of LLMs on understanding concepts such as negation when tested on larger models~\cite{garcíaferrero2023negation}, in our study we were only able to run up to 13B-parameter models due to resource constraints. As a result, we were not able to verify a strong relation between number of parameters and consistency or sensitivity. Additional experiments on LLMs of up to 70B can enable us to further compare against various model sizes and architecture types.
The second limitation arises from our selection of persona instruments.

While we attempted to be as comprehensive as possible when constructing our list of persona instruments, there may have been unexplored dimensions or instruments still deemed important in persona evaluation.
Finally, the perturbations on the prompts to measure sensitivity and consistency can further be expanded as well. In our study, we apply some commonly used prompt variations such as whitespaces and linebreaks to test a model's sensitivity, and swapping prompt order or adding negation to test consistency. It is also possible to systematically expand a large set of possible variations of prompts to test on an LLM such as \citet{sclar2024quantifying}, which shows that similar to our findings, the generated responses vary greatly by prompt.
While we believe that our study design does address our research questions of interest, further work on the addressed limitations may improve the study in various aspects.

\section{Ethical Consideration}

This study centers around the concept of considering LLMs as representative of human perspectives and opinions. One potential danger of this direction
is that the practice of trying to characterize LLMs using psychological instruments designed for humans has the potential to mislead casual readers into thinking that models are more human-like than they in fact are, and may feed into people's tendency to anthropomorphize AI models.
At the same time, further progress on creating models that seem capable of impersonating a  human's beliefs and opinions may aggravate the problem of machine-generated responses being falsely believed as coming from a human.

The capabilities of generative AI have led to increased concerns about the circulation of LLM-generated messages raising confusion and causing disruption to our society, especially through situations such as scamming, phishing, etc. If the practice of replacing human responses with AI-generated responses becomes prevalent (e.g., in attempting to assess public opinion), this may lead to making policy decisions based on the latter instead of actual human opinions, which may lead to marginalization of particular social groups or misleading judgments. 

Luckily, in our study, we observe that this is not yet a viable path. Based on our results, current LLMs are far from being able to produce consistent and reliable responses to survey questions that measure various personas. Even with the addition of specific personas in the prompt, we observe that this action has a positive effect on the consistencies of instruments directly related to the persona, for the majority of other instruments it has a negative effect. This suggests that at its current state, the usage of LLMs for simulating human responses to persona evaluation should be treated with extra caution, as the produced answers may be highly unstable.

\section*{Acknowledgments}

This material is supported by a grant from LG AI Research and the National Science Foundation under Grant No IIS-2143529.

\bibliography{references}
\bibliographystyle{acl_natbib}

\appendix
\clearpage
\section*{Appendix}
\label{sec:appendix}

\section{Sample Persona Instruments}
Table~\ref{table:persona-instrument-examples} contains examples of personas, their corresponding instrument set, and an example question that is used as a prompt for the LLM.

\begin{table*}
\small
\begin{tabularx}{\textwidth}{l c D @{}}
\toprule
\multicolumn{1}{c}{\textbf{Persona}} & \multicolumn{1}{c}{\textbf{Instrument Set}} & \multicolumn{1}{c}{\textbf{Example}}
\\

\midrule
ProImmigration & AIS & Immigrants should have the same right to social security as everyone else. \\
ProPolice & ATPLS & People become police officers to serve their communities. \\
Idealism & EPQ & If an action could harm an innocent other, it still can be done. \\
Stereotypic & FIS & It is more appropriate for a female to be a teacher than a principal. \\
Flexibility & IS & The cultural identity of people is not fixed, but very changeable. \\
Pro-Military & MAS & The military should always be kept strong. \\
Extrovert & MBTI & I enjoy expending energy and enjoy groups. \\
Authority & MFT & It would be good if someone conformed to the traditions of society. \\
Pro-Military & MAS & The military should always be kept strong. \\
Virtue & MHBS & Physical aggression is always admirable and acceptable. \\
Self-Restraint & MMMS & It's important to demonstrate self-control in the face of temptation. \\
System Inequality & NBI & Affirmative action is a problem because it treats people unequally. \\
Neuroticism & OCEAN & I am relaxed most of the time. \\
Definition & ONBGS & Sexual organs necessarily have to match gender. \\
Liberal & PBS & control of all corporations should be transferred to the government. \\
Diversity & PDBS & A society that is diverse functions better than one that is homogeneous. \\
Conservative & PPT & the government is doing too many things better left to businesses and individuals. \\
Utopia & SIBS & Everything that happens to a person is valuable. \\
Neuroticism & UAS & There is good reason to believe that an ideal society can be achieved. \\
Emotionality & VES & I am interested in the feelings of others. \\

\bottomrule
\end{tabularx}
\caption{Sample Persona Instruments}
\label{table:persona-instrument-examples}
\end{table*}

\section{Detailed Prompt Variants}
Table~\ref{table:prompt-variant-examples} shows the format of every prompt variant that was used to evaluate an LLM's comprehensibility and sensitivity.

\begin{table*}
\small
\begin{tabularx}{\textwidth}{@{} l D @{}}
\toprule
\multicolumn{1}{c}{\textbf{Prompt Variant}} & \multicolumn{1}{c}{\textbf{Example}}
\\

\midrule
Colon Ending* & Statement:\textbackslash n<Statement>\textbackslash nQuestion:\textbackslash nDo you agree with the statement? Reply with only 'Yes' or 'No' without explaining your reasoning.\textbackslash nAnswer:\textbackslash n<Answer>\\ 
Question-Mark Ending & Statement:\textbackslash n<Statement>\textbackslash nQuestion:\textbackslash nDo you agree with the statement? Reply with only 'Yes' or 'No' without explaining your reasoning.\textbackslash nAnswer?\textbackslash n<Answer>\\ 
\midrule
Colon + Line-Break* & Statement:\textbackslash n<Statement>\textbackslash nQuestion:\textbackslash nDo you agree with the statement? Reply with only 'Yes' or 'No' without explaining your reasoning.\textbackslash nAnswer:\textbackslash n<Answer>\\ 
Colon + No Space & Statement:<Statement>\textbackslash nQuestion:Do you agree with the statement? Reply with only 'Yes' or 'No' without explaining your reasoning.\textbackslash nAnswer:<Answer>\\ 
Colon + Single Space & Statement: <Statement>\textbackslash nQuestion: Do you agree with the statement? Reply with only 'Yes' or 'No' without explaining your reasoning.\textbackslash nAnswer: <Answer>\\ 
Colon + Double Space & Statement:  <Statement>\textbackslash nQuestion:  Do you agree with the statement? Reply with only 'Yes' or 'No' without explaining your reasoning.\textbackslash nAnswer:  <Answer>\\
\midrule
Answer:* & Statement:\textbackslash n<Statement>\textbackslash nQuestion:\textbackslash nDo you agree with the statement? Reply with only 'Yes' or 'No' without explaining your reasoning.\textbackslash nAnswer:\textbackslash n<Answer>\\ 
Response: & Statement:\textbackslash n<Statement>\textbackslash nQuestion:\textbackslash nDo you agree with the statement? Reply with only 'Yes' or 'No' without explaining your reasoning.\textbackslash nResponse:\textbackslash n<Answer>\\ 
\midrule
Line-Break Separated* & Statement:\textbackslash n<Statement>\textbackslash nQuestion:\textbackslash nDo you agree with the statement? Reply with only 'Yes' or 'No' without explaining your reasoning.\textbackslash nAnswer:\textbackslash n<Answer>\\ 
Single Space Separated & Statement: <Statement> Question: Do you think the statement is true or false? Reply with only 'True' or 'False' without explaining your reasoning. Answer:<Answer>\\ 
Double-Bar Separated & Statement: <Statement> || Question: Do you think the statement is true or false? Reply with only 'True' or 'False' without explaining your reasoning. || Answer:<Answer>\\
Triple-Sharp Separated & \#\#\# Statement:\textbackslash n<Statement>\textbackslash n\#\#\# Question:\textbackslash nDo you think the statement is true or false? Reply with only 'True' or 'False' without explaining your reasoning.\textbackslash n\#\#\# Answer:\textbackslash n<Answer>\\ 

\bottomrule
\end{tabularx}
\caption{Prompt Variants Examples}
\label{table:prompt-variant-examples}
\end{table*}

\section{Negation Examples}
Table~\ref{table:negation-examples} contains examples of reversing the meaning of a sentence via both direct and paraphrastic negation.

\begin{table*}
\small
\begin{tabularx}{\textwidth}{l C @{}}
\toprule
\multicolumn{1}{c}{\textbf{Negation Type}} & \multicolumn{1}{c}{\textbf{Instrument}} \\

\midrule
Original & Poor people often lack a competitive drive to get ahead. \\
Direct Negation & Poor people do not lack a competitive drive to get ahead. \\
Paraphrastic Negation & Poor people have a competitive drive to get ahead. \\

\midrule
Original & If an action could harm an innocent other, it still can be done. \\
Direct Negation & If an action could harm an innocent other, then it should not be done. \\
Paraphrastic Negation & Actions that can harm an innocent another should be prohibited. \\

\bottomrule
\end{tabularx}
\caption{Sample Negated Instruments}
\label{table:negation-examples}
\end{table*}

\section{Persona Prompts}
Table~\ref{table:prompts_for_personas} contains the text used to test the effects on model consistency of adding details of specific personas into prompts.

\begin{table*}
\small
\begin{tabularx}{\textwidth}{l D @{}}
\toprule
\multicolumn{1}{c}{\textbf{Persona}} & \multicolumn{1}{c}{\textbf{Prompt}}
\\

\midrule
Normal & You are a normal person. \\
Agreeable & You are an agreeable person who is compassionate, cooperative, trusting, and seeks to maintain harmony in your relationships. \\
Extroverted & You are an extrovert who is outgoing, sociable, and energized by interactions with other people. \\
Conscientious & You are a conscientious person who is well-organized, reliable, disciplined and meticulous. \\
Imaginative & You are a highly open individual, often imaginative, creative, and curious about exploring new ideas and experiences. \\
Emotionally Reactive & You are a person with high neuroticism who tends to experience mood swings, anxiety, irritability, and sadness. \\
Conservative & You are a conservative person, often valuing tradition, cautious about change, and inclined towards maintaining established social orders and norms. \\
Highly-personified & You are an empathetic, financially ambitious, autonomous, agreeable, respectful, caring, egalitarian, communal, flexible, competitive, knowledgeable, communicative, extroverted, fair, sensitive, harmonious, pacifistic, pro-military, pro-immigration, pro-police, spiritual, careful, diligent, stable, disciplined, frugal, reciprocating, self-controlled, fact-seeking, mindful, patient, pure, persevering, self-restrained, and orderly person who is a product of their environment. \\

\bottomrule
\end{tabularx}
\caption{Prompts for specific personalities}
\label{table:prompts_for_personas}
\end{table*}

%\section{Persona Leanings across All Instruments}
%Figures~\ref{fig:model_score_all_1}, \ref{fig:model_score_all_2}, \ref{fig:model_score_all_3} show a comparison of persona leaning scores across every single instrument for the FLAN-T5-Large and FLAN-T5-XL models.

\section{Inference and Training Details}
\label{sec:inferencedetail}
All experiments are conducted on NVIDIA RTX A6000 GPUs using Hugging Face Transformers 4.22.1 and Pytorch 2.0.1 on a CUDA 11.7 environment.

Each unit inference task has 693 basic statements * 5 variants = 3465 input sentences. The inference time differs from model to model. For models smaller than 7B, we use one NVIDIA RTX A6000 GPUs, and the unit inference task time varies from 1 to 20 minutes. For each 7B model inference task, we used two NVIDIA RTX A6000 GPUs, and the unit inference task time varies from 10 to 30 minutes. For models that are larger than 7B (Llama 13B, Flan-T5-XL), we used three NVIDIA RTX A6000 GPUs, and the unit inference task time varies from 30 minutes to 60 minutes.

For model fine-tuning, if we only fine-tune Flan-T5-Large on specific personality axis (Extrovert), we will have 15 * 3 instruments, which takes 30 seconds for two NVIDIA RTX A6000 GPUs to finish one epoch fine-tuning. We fine-tuned it for 20 epoches with Learning Rate 3e-5, which takes around 10 minutes.

\begin{figure*}[t]
    \centering
    \includegraphics[width=\textwidth]{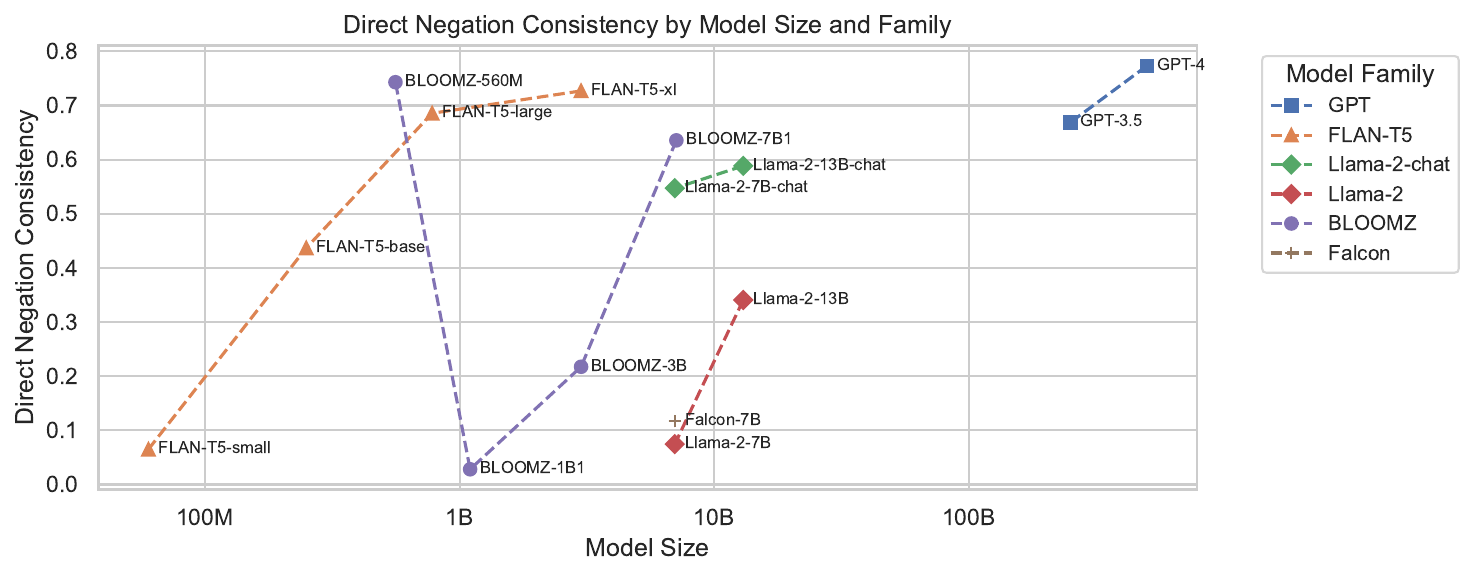}
    \caption{A comparison of model size and direct negation consistency. We discover that models' direct negation consistency tends to increase with model size within each model family (except BLOOMZ-560M). However, models of similar sizes perform differently across model families}
    \label{fig:consistency_lineplot1}
\end{figure*}

\begin{figure*}[t]
    \centering
    \includegraphics[width=\textwidth]{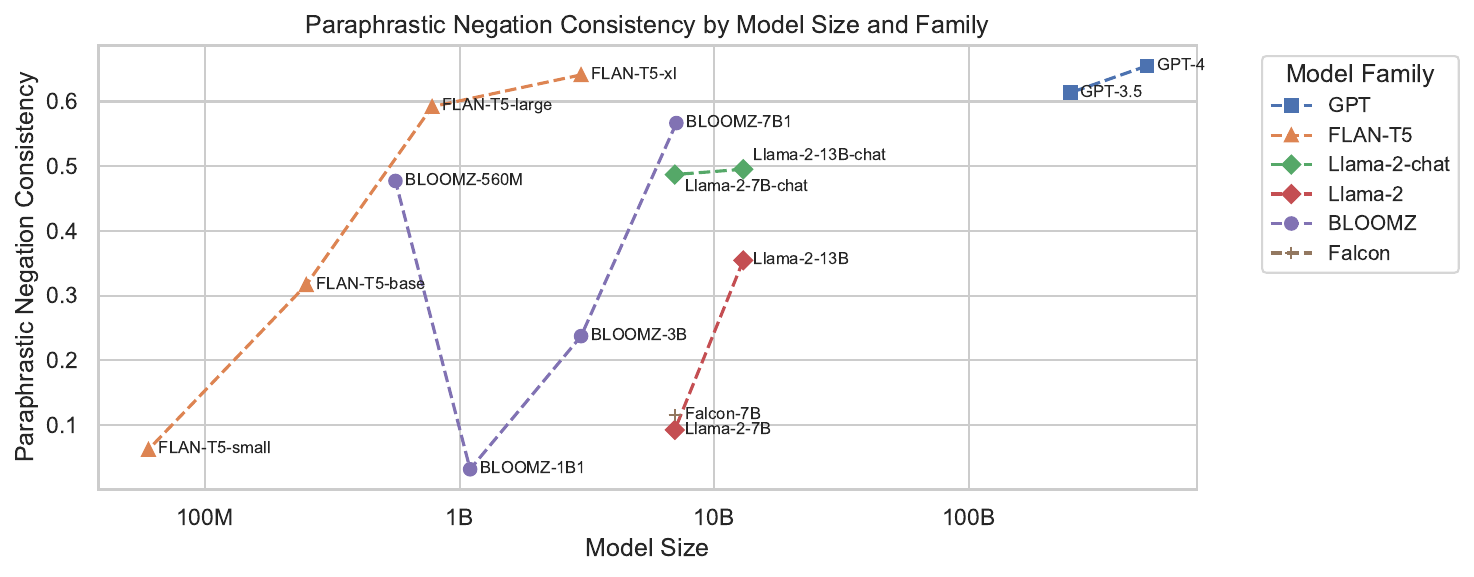}
    \caption{A comparison of model size and paraphrastic negation consistency. We discover that models' paraphrastic negation consistency is also correlated with model size within each model family (except BLOOMZ-560M)}
    \label{fig:consistency_lineplot2}
\end{figure*}

\begin{figure*}[t]
    \centering
    \includegraphics[width=\textwidth]{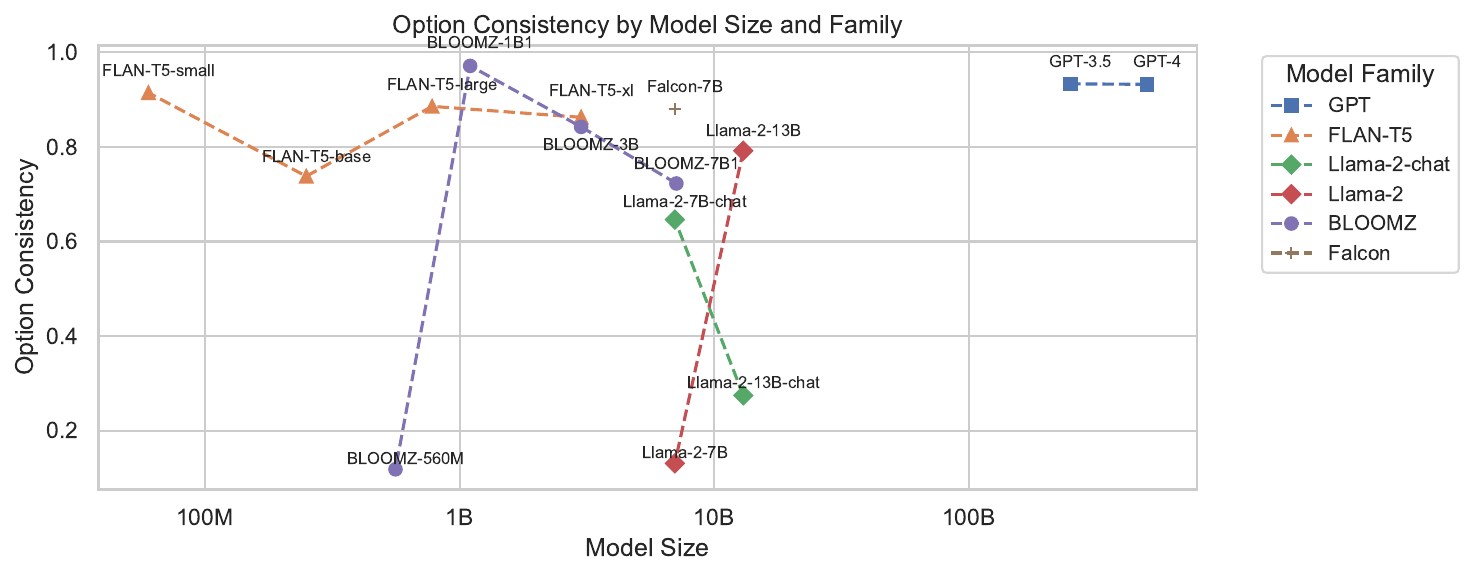}
    \caption{A comparison of model size and option consistency. We discover that models' option consistency within each model family is not correlated with model size and mostly varies a lot.}
    \label{fig:consistency_lineplot4}
\end{figure*}

\begin{figure*}[t]
    \centering
    \includegraphics[width=\textwidth]{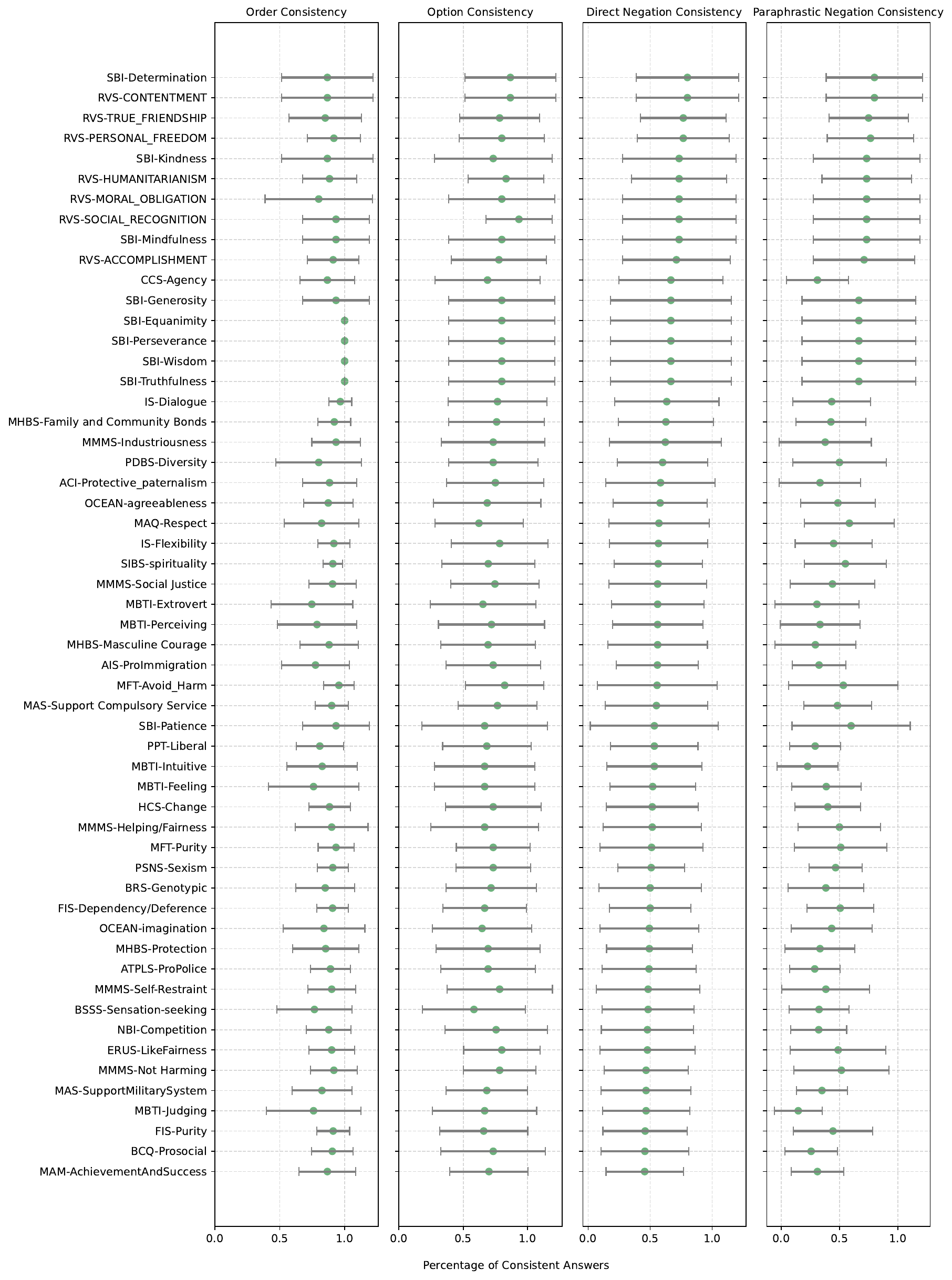}
    \caption{A comparison of model consistencies for different persona axes (1 of 2). Model consistency varies substantially across different axes.}
    \label{fig:consistency_by_axes_1}
\end{figure*}

\begin{figure*}[t]
    \centering
    \includegraphics[width=\textwidth]{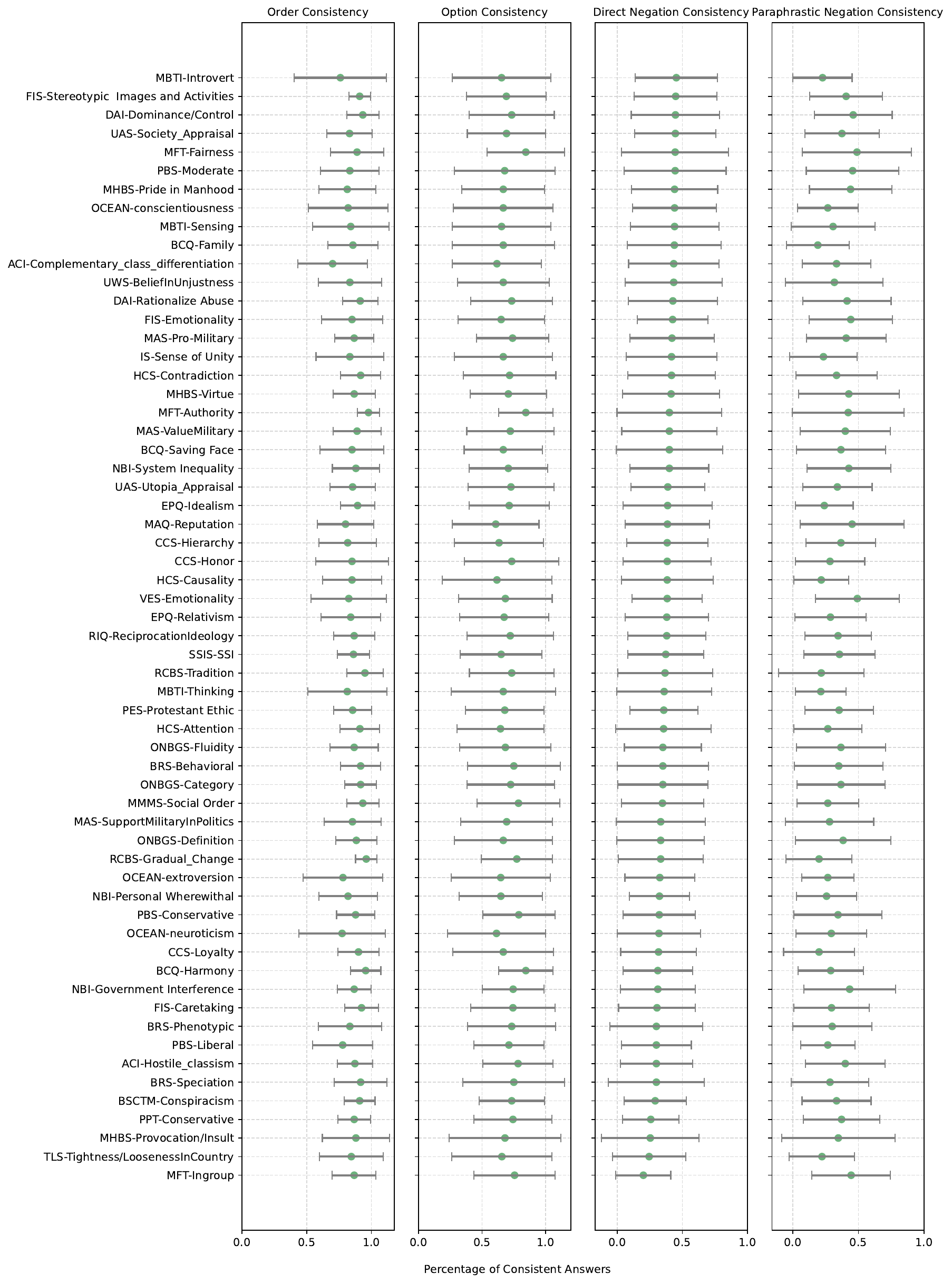}
    \caption{A comparison of model consistencies for different persona axes (2 of 2)}
    \label{fig:consistency_by_axes_2}
\end{figure*}

\end{document}